\documentclass[10pt,twocolumn,letterpaper]{article}

\usepackage[pagenumbers]{cvpr} %

\usepackage{arydshln}
\usepackage{multirow}
\usepackage{algorithm}      %
\usepackage{algpseudocode}  %
\definecolor{cvprblue}{rgb}{0.21,0.49,0.74}
\usepackage[pagebackref,breaklinks,colorlinks,allcolors=cvprblue]{hyperref}
\usepackage{tikz}
\usetikzlibrary{
    arrows.meta,
    positioning,
    fit,
    backgrounds,
    calc
}
\def\confName{CVPR}
\def\confYear{2027}

\title{Plan Big, Paint Small: Semantic-Aware Computation Allocation for Efficient Autoregressive Video Generation}
\newcommand{\method}{\textsc{BigSmall}}
\newcommand{\methodF}{\textsc{BigSmall}-Fast}

\author{Yixuan Luo\\
University of Chicago\\
\and
Yanjing Li\\
Northeastern University\\
}

\begin{document}
\maketitle
\begin{abstract}
Autoregressive (AR) video diffusion streams a video chunk by chunk, but large AR models are too slow for real-time use. We ask where in an AR rollout a large model is actually needed. Pairing Krea-Realtime-14B with Self-Forcing-1.3B, which share their latent space and denoising schedule, we find that the order of the two models matters: when the 14B model generates only the first chunk and the 1.3B model continues, the rollout keeps 95\% of the 14B model's semantic advantage, whereas the reverse allocation keeps 16\% at about ten times the cost. The large model decides what the video shows; the small model only has to carry it forward. We turn this into \method{}, a training-free planner--renderer sampler. The 14B model plans the opening chunk and sketches the chunks in which content changes, which preserves motion, and the 1.3B model renders the rest; a faster variant, \methodF{}, plans only the opening chunk. A cross-model KV cache that is approximate, windowed and lazy lets the planner leave and re-enter the rollout at low cost. On VBench, \method{} is $3.02\times$ faster than the 14B model at an unchanged semantic score and motion, and $1.55\times$ faster than SDVG, which lets the small model draft first; \methodF{} reaches $3.77\times$ at the same semantic score, trading away 10\% of the motion. On minute-long videos (LoCoT2V-Bench), where the 14B model degrades when it reads its own long history, \method{} is $3.14\times$ faster than the 14B model and scores higher ($+1.78$).
\end{abstract}

\section{Introduction}
\label{sec:intro}

Autoregressive (AR) video diffusion generates a video chunk by chunk, each
chunk conditioned on the previous ones through a key-value (KV) cache
\cite{chen2024diffusionforcing,sandai2025magi,chen2025skyreels,zhang2025framepack,jin2025pyramidal}.
Frames can be shown as soon as they are produced, and the user can steer the
video while it is generated, for example by changing the prompt. Distilled
few-step causal generators make this practical
\cite{yin2025causvid,huang2025selfforcing,yin2024dmd,yin2024dmd2}, and, as for
bidirectional video models \cite{ho2022video,blattmann2023svd,yang2025cogvideox,kong2024hunyuanvideo,wan2025wan},
scaling them improves quality: Krea-Realtime-14B \cite{krea2025} follows
prompts better than the 1.3B Self-Forcing model, but runs at 3.3 instead of
17.1 frames per second. \textbf{Can we keep what the large model adds without
paying for it on every chunk?}

\begin{figure}[t]
  \centering
  \includegraphics[width=\linewidth]{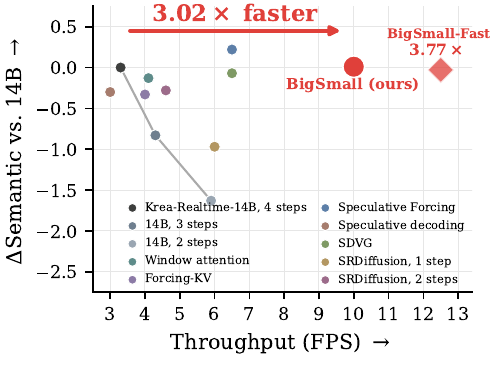}
\caption{\textbf{Semantic score versus throughput on VBench} (946 prompts,
5 seeds). \method{} is $3.02\times$ faster than Krea-Realtime-14B at the same
semantic score, and \methodF{}, which calls the 14B model only for the first
chunk, is $3.77\times$ faster, also at the same semantic score. Fewer 14B steps
lose semantics quickly; single-model accelerators and large--small
cooperation stay at or below 6.5 FPS. $\Delta$Semantic: paired difference to
the 14B model; FPS: DiT throughput for 81-frame videos on one GPU
(Tab.~\ref{tab:main_results}).}
  \label{fig:teaser}
\end{figure}

A natural answer is to let a large and a small model cooperate, as in
language-model decoding \cite{leviathan2023speculative,chen2023speculative,kim2023bild}
and image diffusion \cite{pan2025tstitch,liu2023omsdpm,balaji2022ediff}. In AR
video, existing schemes either let the small model draft every chunk and
verify it, with the large model or a reward model \cite{hu2026sdvg}, or split
the denoising steps of every chunk between the models
\cite{cheng2025srdiffusion}. Both are costly: exact speculative decoding is
slower than the 14B model ($0.90\times$), SDVG reaches $1.95\times$, and an
SRDiffusion-style cascade needs two 14B steps per chunk ($1.40\times$).
Single-model accelerations
\cite{ma2024deepcache,liu2024timestep,zhao2025pab,xi2025sparsevideogen,ji2026forcingkv,choi2026speculativeforcing}
still run the large model on every chunk and are complementary. SDVG already
generates its first block with the large model, but still drafts, decodes and
scores every later chunk. We instead decide along video time, \emph{in
advance}, which chunks need the large model at all.

We show that the large model should go \emph{first}. Generating only the
first chunk with the 14B model keeps 95\% of its semantic advantage on VBench
\cite{huang2024vbench}; the reverse, a 1.3B first chunk followed by six 14B
chunks, keeps 16\% at about ten times the cost (Sec.~\ref{sec:analysis}). The
chunk that introduces the content decides what the video shows, and the small
model carries it forward. Inside this chunk the large model must take most
denoising steps; on later chunks a single large-model step fixes the layout
and keeps the motion that the small model alone would damp. Over minute-long videos, the 14B model fades when it reads its own
long history, while the self-forced 1.3B model stays stable
(Sec.~\ref{sec:locot2v}).

Turning this into a sampler raises a problem specific to AR generation:
each model conditions on its own KV cache, and while the large model is idle
its cache falls behind the video. Rebuilding it exactly costs one 14B forward
per chunk; image cascades \cite{ho2022cascaded}, which exchange only latents,
never face this. Our \textbf{cross-model KV cache} keeps the renderer's cache
exact and makes the planner's \emph{approximate} (only moving tokens are
re-encoded), \emph{windowed} (a sink and the previous chunk, as in
\cite{xiao2024streamingllm}) and \emph{lazy} (built only when needed). This
suffices because the planner's prediction is re-noised, and only its coarse
structure survives.

\method{} (Fig.~\ref{fig:method}) uses the 14B model to plan the first chunk
and to sketch the first denoising step of chunks in which a zero-cost change
score signals new content; the 1.3B model renders the rest, and \methodF{}
keeps only the plan chunk. Unlike long-video and real-time AR models trained
for speed or length \cite{yang2025longlive,krea2025}, \method{} needs no
training and combines two existing models. \method{} is $3.02\times$ faster
than Krea-Realtime-14B on VBench at an unchanged semantic score and motion,
and $1.55\times$ faster than SDVG; \methodF{} reaches $3.77\times$. On
minute-long videos from LoCoT2V-Bench
\cite{zheng2026locot2vbenchbenchmarkinglongformcomplex}, \method{} is
$3.14\times$ faster than the 14B model and scores higher ($+1.78$). Our
contributions are:
\begin{itemize}
\item \textbf{Large first, small after.} Along video time, the large model
  should generate the chunk that introduces the content and the small model
  the rest; this holds for three independently trained renderers, and
  small-first schemes are slower at equal quality. Unlike prior cooperation,
  which splits the denoising steps of every chunk or verifies every draft, we
  decide in advance whether a chunk needs the large model, and explain why the
  first chunk matters.
\item \textbf{A division of labour.} The plan chunk sets the semantics,
  single-step sketches keep the motion, and the small model renders the rest
  and, over a minute, is the more stable renderer.
\item \textbf{\method{} with a cross-model KV cache.} An approximate, windowed
  and lazy planner cache lets the large model be absent from most chunks; two
  training-free operating points, \method{} ($3.02\times$) and \methodF{}
  ($3.77\times$), keep the 14B model's semantics.
\end{itemize}

\section{Related Work}
\label{sec:related_work}

\subsection{Autoregressive Video Generation and Efficient Inference}

Autoregressive video diffusion enables streaming and long-form
generation. Diffusion Forcing combines causal sequence modeling with
diffusion \cite{chen2024diffusionforcing}, CausVid and Self Forcing
build few-step autoregressive generators with improved efficiency and
rollout consistency \cite{yin2025causvid,huang2025selfforcing}, and
recent works target real-time interactive and long-form generation
\cite{yang2025longlive,krea2025}. These works train a single model;
\method{} is training-free and combines two existing models, and we
find that over a minute the smaller self-forced model is the more
stable of the two.

Other methods reduce inference cost via feature caching, sparse
attention, and KV cache compression
\cite{liu2024timestep,li2025radial,xi2025sparsevideogen,
ji2026forcingkv,ma2026flowcache}, or adapt speculative decoding with
draft-and-verify strategies \cite{leviathan2023speculative,hu2026sdvg,
choi2026speculativeforcing}. Single-model accelerations make the large
model cheaper but still run it on every chunk, and are complementary
to ours. KV cache compression shortens context to save computation;
we find a short context also suits planning (Sec.~\ref{sec:context}),
while long-range state is better carried by the drift-free small model.
Speculative schemes let the small model go first, but since a whole
chunk must be verified, drafts are rarely accepted
(Sec.~\ref{sec:analysis}). \method{} reverses the order: the large
model introduces the content, and the small model continues unverified.

\subsection{Large and Small Model Cooperation}

Large--small cooperation has been explored to accelerate diffusion
sampling. T-Stitch replaces part of a large model's trajectory with a
smaller model \cite{pan2025tstitch}, SRDiffusion assigns sketching and
rendering stages to models of different capacities
\cite{cheng2025srdiffusion}, and TRACK routes models by trajectory
disagreement without training \cite{munir2026track}. These methods
allocate models only along the denoising trajectory, still calling
the large model on every image or chunk. AR rollout adds a second
axis, video time, which we find matters more: the first chunk decides
the semantics. They also exchange only latents, whereas AR models
condition on a KV cache, which \method{} shares across the two models.
\section{Where Does Scale Matter in an AR Rollout?}
\label{sec:analysis}

\paragraph{Setup.}
\label{sec:setup}
We pair Krea-Realtime-14B~\cite{krea2025} with
Self-Forcing-1.3B~\cite{huang2025selfforcing}. Both are block-causal diffusion
transformers in the same latent space; an 81-frame video has seven chunks of
three latent frames (4,680 tokens each), and both models share the schedule
$t \in \{1000, 937.5, 833.3, 625\}$. After denoising step $k$, the predicted
clean latent is re-noised as
\begin{equation}
  x^{(k+1)} = (1-\sigma_{k+1})\,\hat{x}_0^{(k)} + \sigma_{k+1}\,\epsilon ,
  \label{eq:renoise}
\end{equation}
and a context pass at $t=0$ writes each finished chunk into the model's KV
cache. Because the two models share latents and schedule, a noisy latent can
be handed from one to the other at any step. Within each paired comparison we
fix the initial and re-noising noise and the prompt, so rollouts differ only in
which model runs which step of which chunk. We evaluate on VBench
\cite{huang2024vbench}
with paired bootstrap 95\% CIs, and report the \emph{gap recovered}: the
fraction of the 1.3B-to-14B semantic gap that a mixed rollout closes.

\paragraph{Why not let the small model draft?}
Draft-and-verify schemes \cite{leviathan2023speculative,hu2026sdvg} let the
small model go first. In AR video, the verification unit is a chunk of 4,680
latent tokens: verifying $K$ drafted chunks costs $K$ additional 14B forwards
(a $3.95\times$ overhead for $K{=}4$), and although maximal coupling accepts
97--99\% of individual elements, a whole chunk-step is accepted with probability
below 0.3\%. The large model is better used to \emph{generate}
where it matters than to check the small model everywhere.

\subsection{Who Begins Decides}
\label{sec:who_begins}

\begin{figure}[t]
  \centering
  \includegraphics[width=\linewidth]{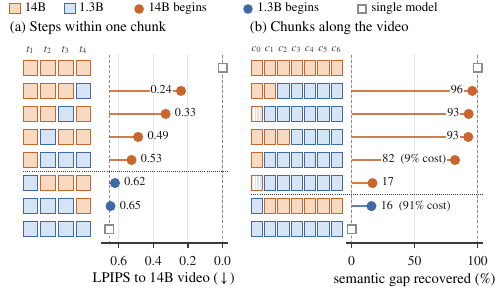}
  \caption{\textbf{Who begins decides.} Noise fixed; only the model allocation
  changes. Each row is an allocation (peach: 14B, blue: 1.3B). (a) Within one
  chunk, over denoising steps $t_1$--$t_4$: LPIPS to the 14B video (24 prompts).
  (b) Along the video, over chunks $c_0$--$c_6$ (a split $c_0$: 14B on its first one
  or three steps): semantic gap recovered on VBench; cost = share of
  the 14B model's additional computation.}
  \label{fig:who_begins}
\end{figure}

\paragraph{Along the video, the first chunk decides.}
\label{sec:video_time}
Letting the 14B model generate only the first chunk and the 1.3B model the
remaining six recovers almost the full semantic score of the 14B model
($-0.03$; 95\% of the gap) with only 9\% of the
14B model's additional computation (Fig.~\ref{fig:who_begins}b); we keep this
allocation as our fastest variant, \methodF{}. The gain is
not confined to the first chunk: chunks 1--6 are generated by the same 1.3B
model from the same noise as in the 1.3B rollout, yet multiple-object accuracy
rises from 85.5 to 92.1 (14B: 92.5). Among the videos in which the 1.3B model
misses an object in most frames, the 14B first chunk alone repairs 6 of 10
(the full 14B rollout: 7 of 10). The reverse allocation shows the asymmetry: a
1.3B first chunk followed by six 14B chunks recovers only 16\% of the gap, is
indistinguishable from the 1.3B model ($+0.39$) and
significantly below the 14B model ($-2.11$), while spending
91\% of the additional computation. Content introduced by the first chunk
persists; content missing from it is rarely added later: objects absent from
the first chunk appear later in only 1--2\% of videos, even when the 14B model
generates chunks 1--6. This agrees with SDVG
\cite{hu2026sdvg}, which accepts 93.8\% of 1.3B drafts and attributes the
large model's systematic contribution to a mandatory first block.

\paragraph{Inside a chunk, the first step fixes the layout.}
\label{sec:denoising_time}
The model that takes the first denoising step determines where things are
(Fig.~\ref{fig:who_begins}a): one 14B first step followed by three 1.3B steps
(bsss) reaches an LPIPS of 0.526 to the 14B video, whereas a 1.3B first step
followed by three 14B steps (sbbb) stays at 0.622, close to the 1.3B video
(0.655). Semantics, however, are committed over the whole first chunk. In a
$2{\times}2$ design on the first chunk, giving step 1 to the 14B
model has no significant effect on the semantic score ($+0.37$), whereas giving it steps 2--4 has a large one ($+1.68$); bbbb, sbbb and bsss recover 82\%, 69\% and 17\% of the gap.
The large model should therefore take all steps of the chunk that introduces
the content, and a single first step suffices only to anchor the layout of
later chunks.
\begin{figure*}[t]
  \centering
    \includegraphics[width=\textwidth]{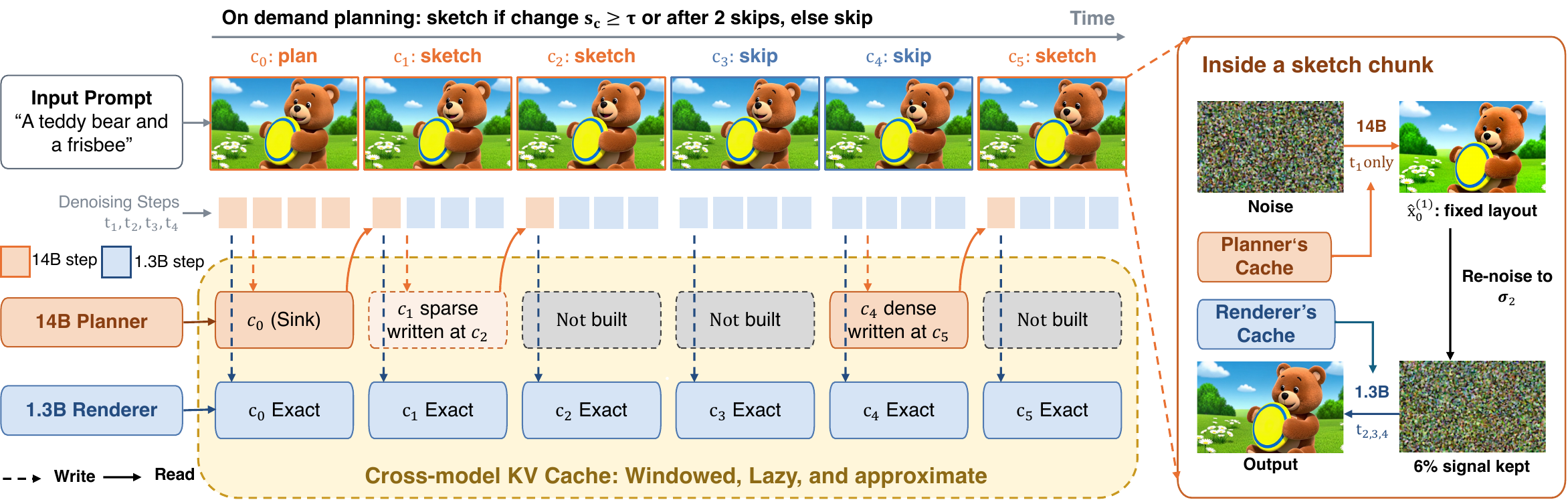}
\caption{\textbf{Overview of \method{}} on a real rollout (``a teddy bear and a
frisbee''). \emph{Top:} a zero-cost change score decides, chunk by chunk,
whether the 14B planner is called. The planner generates the plan chunk $c_0$
with all four denoising steps and only the first step $t_1$ of each sketch
chunk; the 1.3B renderer takes all other steps (squares: denoising steps
$t_1$--$t_4$; orange: 14B, blue: 1.3B). \emph{Bottom:} each model reads its own
KV cache; dashed arrows write a finished chunk into a cache, solid arrows read
from it. The renderer's cache is exact: every chunk is written. The planner's
cache is \emph{windowed}: a planner call reads only the sink $c_0$ and the
previous chunk. It is \emph{lazy}: that previous chunk is written only when the
planner needs it ($c_1$ when $c_2$ is sketched, $c_4$ when $c_5$ is sketched),
and other chunks are never written. It is \emph{approximate}: a chunk is
re-encoded sparsely, only its moving tokens, while the others copy K/V from the
previous frame with re-rotated RoPE; it is re-encoded densely if the chunk
before it was never written, as for $c_4$. \emph{Right:} inside sketch chunk
$c_5$, the planner's first step from noise already fixes the layout
($\hat{x}_0^{(1)}$); it is re-noised to $\sigma_2$, keeping 6\% of the signal, and
the renderer completes the chunk from its own cache.}
  \label{fig:method}
\end{figure*}
\subsection{What Does the Planner Need to See?}
\label{sec:context}

A planner that only fixes the layout of a chunk needs little context. Its
first-step prediction is re-noised to $\sigma_2=0.9375$ before the renderer
continues, so only 6.25\% of the predicted signal survives
(Eq.~\ref{eq:renoise}); low-pass filtering the planner's context by $4\times$
changes the result only slightly (LPIPS 0.535 vs.\ 0.544). Long
context can even hurt. After a prompt switch, both models keep generating the
old scene: the semantic score of the new prompt drops from about 85 to 35.6 for
the 1.3B model and to 23.7 for the 14B model, which is significantly worse
($-11.35$). The larger model relies more on its context,
and its context shows the old scene. An exact 14B cache would cost one extra
14B forward per chunk; a short, approximate planner context is both cheaper
and better suited to planning.

\paragraph{Implications.}
(i)~The 14B model should generate the chunk that introduces the content with
all its steps; (ii)~on later chunks, one 14B step suffices to anchor the layout
when content changes; (iii)~the 1.3B model renders everything else and keeps
the exact cache; (iv)~the planner's context can be short, approximate and
built only when needed. For a short clip with one prompt, most content enters
in the first chunk; for long videos and changing prompts, planning must be
re-activated on demand.

\section{\method: Plan Big, Paint Small}
\label{sec:method}

Sec.~\ref{sec:analysis} implies a schedule in which the 14B model $B$
generates where content is introduced and the 1.3B model $S$ paints the rest.
\method{} implements it without training (Fig.~\ref{fig:method}). At step $k$
of chunk $c$, a model $M \in \{B, S\}$ predicts
$\hat{x}_0^{(k)} = M(x_c^{(k)}, t_k \mid \mathcal{C}_M)$ from its own KV cache
$\mathcal{C}_M$, and the prediction is re-noised to $t_{k+1}$ by
Eq.~\eqref{eq:renoise}, so the latent can pass between the models at any step.

\subsection{Per-chunk schedule}
\label{sec:schedule}
Each chunk $c$ receives a planner budget $F_c \in \{0, 1, 4\}$, the number of
its denoising steps taken by the planner:
\begin{equation}
  \hat{x}_0^{(k)} =
  \begin{cases}
    B(x_c^{(k)}, t_k \mid \mathcal{C}_B), & k \le F_c, \\
    S(x_c^{(k)}, t_k \mid \mathcal{C}_S), & k > F_c.
  \end{cases}
  \label{eq:schedule}
\end{equation}
The first chunk is a \emph{plan} chunk ($F_0 = 4$): it commits the semantics
of the video, which takes all of its steps (Sec.~\ref{sec:denoising_time}). A
later chunk that may introduce new content is a \emph{sketch} chunk
($F_c = 1$): the planner's first step anchors its layout and the renderer
refines it. All other chunks are \emph{skip} chunks ($F_c = 0$). The planner's
cost is then proportional to the number of chunks on which it is called,
provided that calling it does not require maintaining its context at every
chunk; this is what the cross-model KV cache provides.

\subsection{Cross-model KV cache}
\label{sec:cross_model_cache}
Each model attends only to its own cache, but both caches must describe the
same video, painted by both models. The renderer's cache $\mathcal{C}_S$ is
exact and dense: $S$ runs a context pass at $t=0$ after every chunk and carries
the long-range consistency of the video. An exact $\mathcal{C}_B$ would cost a
second 14B forward per chunk, yet it only supports a coarse prediction that is
re-noised to $\sigma_2 = 0.9375$ (Sec.~\ref{sec:context}). We therefore build
$\mathcal{C}_B$ in three cheap ways.

\paragraph{Approximate.}
The planner's context pass re-encodes only moving tokens: the top 25\% of
tokens of each latent frame ranked by $|\hat{x}^{f}-\hat{x}^{f-1}|$, dilated by
$3{\times}3$ ($\approx$36\% of tokens). Every other token inherits the key and
value at the same position from the newest frame already in $\mathcal{C}_B$.
Keys are cached after the rotary embedding, so we re-position them exactly by
the temporal RoPE rotation for the frame offset $\Delta f$; values carry no
position and are copied. All entries are computed from clean latents, as in
the $t{=}0$ context pass of Self-Forcing, so no noise-level correction is
needed.

\paragraph{Windowed.}
$B$ attends only to a sink frame and to chunk $c{-}1$. This is not only a cost
measure. Conditioned on its own long rolling history, $B$ degrades: over a
minute its frames fade and its content freezes, and after a prompt switch it
keeps the old scene (Secs.~\ref{sec:context} and~\ref{sec:locot2v}). Long-range
consistency is therefore left to $\mathcal{C}_S$; $S$ was trained on its own
rollouts \cite{huang2025selfforcing} and stays stable over long horizons, so
$\mathcal{C}_B$ neither grows with the video nor accumulates $B$'s drift.

\paragraph{Lazy.}
Because of the window, the context pass of chunk $c{-}1$ into $\mathcal{C}_B$
is deferred until chunk $c$ calls the planner; it is dense if chunk $c{-}2$ was
not encoded either. Chunks that are not followed by a planner call are never
encoded: leaving the rollout is free, and re-entering costs one short context
pass.

\subsection{On-demand planning}
\label{sec:on_demand}
Whether a chunk $c \ge 2$ is a sketch or a skip chunk is decided by a
zero-cost change score between the two previous clean chunks. Each token of
$\hat{x}_{c-1}$ is matched to its nearest counterpart within $\pm 2$ positions
in $\hat{x}_{c-2}$, and the residuals $r$ give
\begin{equation}
  s_c = \operatorname{mean}\big(\operatorname{MaxPool}_{3\times3}(r)\big).
  \label{eq:change_score}
\end{equation}
Chunk $c$ is sketched if $s_c \ge \tau$, with $\tau$ calibrated on held-out
prompts for a target skip rate of 70\%; chunk~1 is always sketched. At most two
consecutive chunks are skipped (\emph{run cap}), since a static stretch can
hide slow motion that the renderer alone would damp, and the first two chunks
after a prompt switch are sketched, since $s_c$ would otherwise compare chunks
generated under different prompts. Orthogonally, renderer tokens whose
prediction changes by less than a calibrated threshold between steps 2 and 3
skip step 4 and keep their cache slots (\emph{late exit}).

\paragraph{Long videos.}
Beyond 81 frames both caches roll: $\mathcal{C}_S$ keeps a sink chunk and the
six most recent chunks, whose keys are re-rotated by the frame offset when the
window shifts, and $\mathcal{C}_B$ is windowed anyway. Neither model's context
grows with the video, and the planner's cost scales with the number of content
changes rather than with the video length.

\begin{table*}[t]
\centering
\small
\setlength{\tabcolsep}{4.2pt}
\caption{\textbf{Standard VBench} (946 extended prompts, five seeds). Time: DiT
time per video. $\Delta$Sem: paired semantic
difference to the 14B model (95\% CI) on the videos both methods share (same prompt and seed), so it can differ slightly from the difference of the Semantic columns, which use all videos. $^\dagger$Plan chunk only: 14B on chunk 0, 1.3B on all others.}
\label{tab:main_results}
\resizebox{\textwidth}{!}{
\begin{tabular}{lccccccccc}
\toprule
Method & Time (s) & FPS & Speed-up & Total & Quality
& Semantic & $\Delta$Sem (95\% CI) & Dynamic & Multi-obj. \\
\midrule
Krea-Realtime-14B
& 24.50 & 3.3 & $1.00\times$ & 84.25 & 84.76 & 82.21
& -- & 74.7 & 92.5 \\
Self-Forcing-1.3B
& 4.74 & 17.1 & $5.17\times$ & 84.07 & 84.83 & 81.01
& $-1.08$ [$-1.84,-0.33$] & 65.8 & 85.5 \\
14B, 2 steps
& 13.84 & 5.9 & $1.77\times$ & 83.77 & 84.32 & 81.56
& $-1.63$ [$-2.82,-0.51$] & 75.0 & 87.6 \\
14B, 3 steps
& 18.74 & 4.3 & $1.31\times$ & 84.14 & 84.59 & 82.35
& $-0.83$ [$-1.70,+0.02$] & 72.2 & 88.9 \\
\midrule
\multicolumn{10}{l}{\textit{Single-model AR acceleration}} \\
Forcing-KV
& 20.21 & 4.0 & $1.21\times$ & 84.26 & 84.86 & 81.88
& $-0.33$ [$-0.54,-0.13$] & 82.8 & 91.4 \\
Speculative Forcing
& 12.40 & 6.5 & $1.98\times$ & 84.16 & 84.60 & 82.41
& $+0.22$ [$-0.12,+0.57$] & 73.3 & 92.3 \\
\midrule
\multicolumn{10}{l}{\textit{Large--small cooperation}} \\
Speculative decoding
& 27.21 & 3.0 & $0.90\times$ & 84.27 & 84.86 & 81.91
& $-0.30$ [$-0.77,+0.18$] & 76.4 & 91.7 \\
SDVG
& 12.54 & 6.5 & $1.95\times$ & 84.37 & 84.92 & 82.16
& $-0.07$ [$-0.32,+0.19$] & 72.2 & 92.1 \\
SRDiffusion, 1 step
& 13.61 & 6.0 & $1.80\times$ & 84.22 & 84.98 & 81.18
& $-0.97$ [$-1.72,-0.35$] & 68.1 & 86.5 \\
SRDiffusion, 2 steps
& 17.56 & 4.6 & $1.40\times$ & 84.31 & 84.92 & 81.89
& $-0.28$ [$-0.81,+0.20$] & 72.2 & 91.0 \\
\hdashline\noalign{\vskip 2pt}
\multicolumn{10}{l}{\textit{Ours}} \\
\method{} w/o plan chunk
& 6.16 & 13.1 & $3.98\times$ & 84.27 & 85.00 & 81.34
& $-0.86$ [$-1.53,-0.18$] & 68.3 & 86.2 \\
\method{} w/o run cap
& 7.48 & 10.8 & $3.28\times$ & 84.18 & 84.68 & 82.21
& $0.00$ [$-0.21,+0.22$] & 70.6 & 92.3 \\
\midrule
\textbf{\methodF{}}$^\dagger$
& \textbf{6.50} & \textbf{12.5} & $\mathbf{3.77\times}$ & \textbf{84.38} & \textbf{84.94} & \textbf{82.15}
& $\mathbf{-0.03}$ [$\mathbf{-0.30},\mathbf{+0.24}$] & \textbf{72.2} & \textbf{92.1} \\
\textbf{\method{}}
& \textbf{8.11} & \textbf{10.0} & $\mathbf{3.02\times}$
& \textbf{84.32} & \textbf{84.86} & \textbf{82.18}
& $\mathbf{+0.01}$ [$\mathbf{-0.22},\mathbf{+0.26}$]
& \textbf{74.2} & \textbf{92.3} \\
\bottomrule
\end{tabular}
}
\end{table*}
\section{Experiments}
\label{sec:experiments}

\subsection{Experimental Setup}
\label{sec:exp_setup}
\begin{figure*}[t]
\centering
\includegraphics[width=\textwidth]{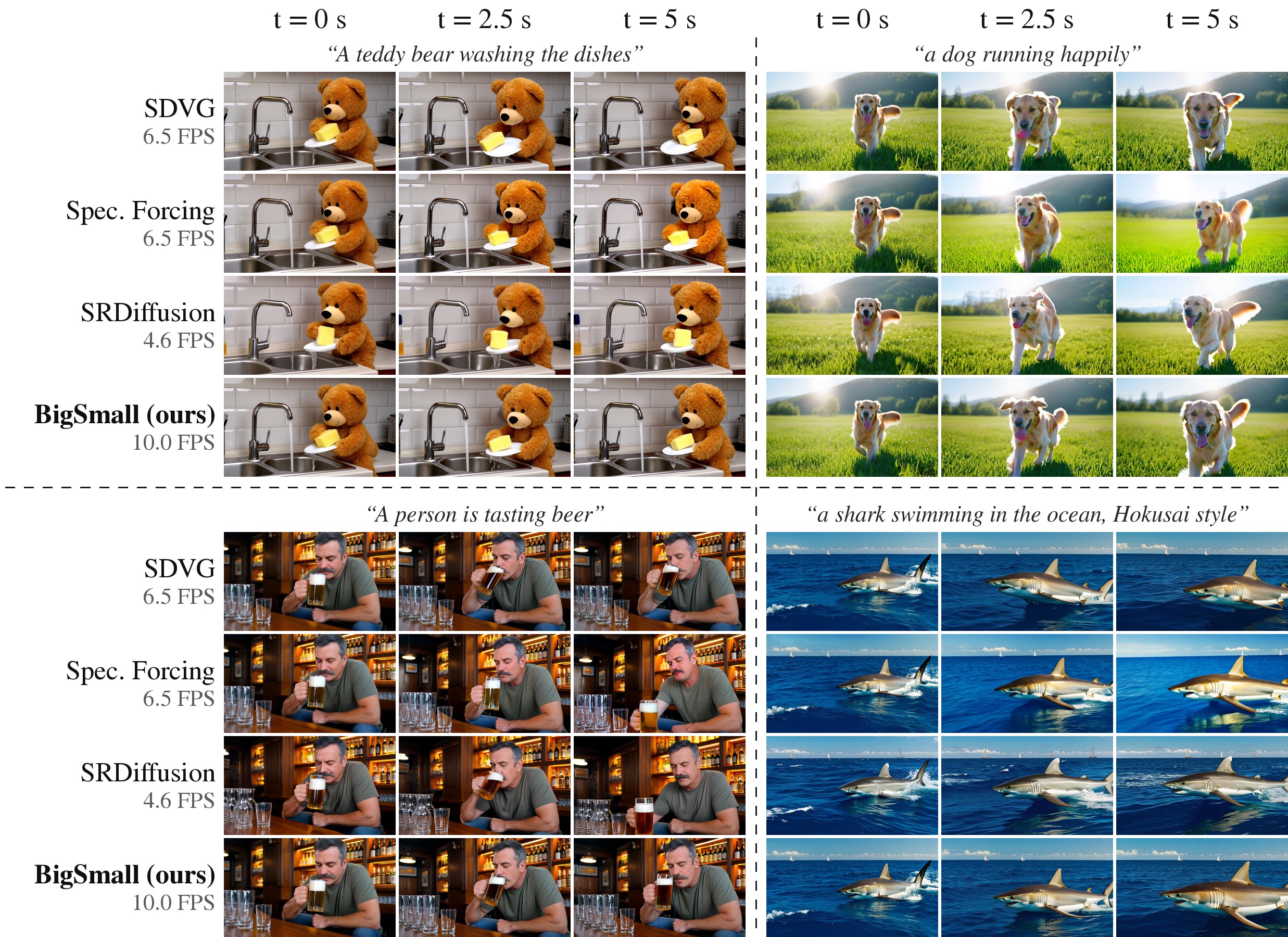}
\caption{\textbf{Qualitative comparison with large--small cooperation and
acceleration baselines} (VBench, seed 0, same noise for all methods; frames at
$t=0$, 2.5 and 5\,s). \method{} produces videos of the same quality as SDVG,
Speculative Forcing and SRDiffusion (two 14B steps per chunk) while running at
10.0 instead of 4.6--6.5 FPS. Prompts drawn at random from the evaluation set and
selected for diverse content and motion.}
\label{fig:vbench_qual}
\end{figure*}
\paragraph{Implementation.}
Planner Krea-Realtime-14B, renderer Self-Forcing-1.3B (DMD), both four-step
Wan2.1-family models; $832\times480$ videos of 81 frames unless noted; BF16,
PyTorch SDPA without compilation, one NVIDIA RTX PRO 6000 Blackwell (96\,GB). We
report the DiT time $T$ per video, excluding VAE decoding, which all methods
share,
and $\mathrm{FPS}=81/T$. All methods share initial and re-noising noise.

\paragraph{Protocols.}
\emph{Standard VBench} \cite{huang2024vbench}: 946 extended prompts released
with Self-Forcing, five seeds, \texttt{vbench\_standard} with official
normalisation, $\mathrm{Total}=(4Q+S)/5$; ablations use seed 0 (944 videos).
\emph{Minute-long video}: the 234 prompts of LoCoT2V-Bench
\cite{zheng2026locot2vbenchbenchmarkinglongformcomplex}, one paragraph per video
(full-prompt input), generated at the reference duration (47\,s on average, up
to 64\,s; 748 frames on average) with a rolling context (sink and six recent
chunks); we report the five official dimensions, perceptual quality (PQ),
text--video alignment (TVA), temporal quality (TQ), HERD and dynamic quality
(DQ), and their mean (Avg); FPS is total frames over total DiT time.
\emph{Prompt switches}: 172 videos of 21 chunks whose prompt switches every
seven chunks. All CIs are paired bootstrap 95\% intervals relative to the 14B
model.

\paragraph{Baselines.}
\emph{Single models}: the 14B model with 4, 3 or 2 steps and the 1.3B model.
\emph{Single-model AR acceleration}: Forcing-KV \cite{ji2026forcingkv} (KV
cache compression) and Speculative Forcing \cite{choi2026speculativeforcing}
(token-level acceleration). \emph{Large--small cooperation}: exact speculative
decoding with maximal coupling; SDVG \cite{hu2026sdvg}, reproduced on the same
model pair with ImageReward-v1.0 verification; and SRDiffusion-style
sketch--render cascades \cite{cheng2025srdiffusion} in which the 14B model
takes the first one or two steps of every chunk with an exact dense cache.
LoCoT2V-Bench uses the single models, Forcing-KV, Speculative Forcing and SDVG.

\subsection{Results on Standard VBench}
\label{sec:standard_vbench}

\begin{table*}[t]
\centering
\caption{\textbf{Minute-long generation} on LoCoT2V-Bench. Avg: official mean of the five
dimensions; $\Delta$Avg: paired difference to the 14B model (95\% CI).
Speed-up from DiT time; SDVG spends another 49\,s per video decoding and
scoring drafts.}
\label{tab:locot2v}
\small
\setlength{\tabcolsep}{5pt}
\begin{tabular}{lcccccccc}
\toprule
Method & Speed-up & PQ & TVA & TQ & HERD & DQ & Avg & $\Delta$Avg (95\% CI) \\
\midrule
Krea-Realtime-14B & $1.00\times$ & 75.65 & 42.41 & 78.98 & 72.79 & 59.31 & 65.83 & -- \\
Self-Forcing-1.3B & $4.83\times$ & 76.20 & 45.25 & 78.41 & 71.28 & 57.45 & 65.72 & $+0.35$ [$-0.56,+1.24$] \\
Forcing-KV & $1.45\times$ & 79.01 & 45.97 & 80.93 & 83.45 & 60.74 & 70.02 & $+4.03$ [$+3.31,+4.74$] \\
Spec.\ Forcing & $1.95\times$ & 67.79 & 42.55 & 78.03 & 61.89 & 58.20 & 61.69 & $-4.15$ [$-5.06,-3.28$] \\
SDVG & $0.91\times$ & 76.47 & 45.87 & 79.88 & 74.30 & 59.31 & 67.17 & $+1.47$ [$+0.73,+2.24$] \\
\midrule
\methodF{} & $4.77\times$ & 75.00 & 47.28 & 80.01 & 70.90 & 57.23 & 66.09 & $+0.77$ [$-0.09,+1.64$] \\
\textbf{\method{}} & $\mathbf{3.14\times}$ & \textbf{77.04} & \textbf{47.02} & \textbf{80.43} & \textbf{74.37} & \textbf{59.86} & \textbf{67.74} & $\mathbf{+1.78}$ [$\mathbf{+0.89},\mathbf{+2.67}$] \\
\bottomrule
\end{tabular}
\end{table*}
\paragraph{Main result.}
\method{} runs at 10.0 FPS, $3.02\times$ faster than Krea-Realtime-14B
(Tab.~\ref{tab:main_results}), with a paired semantic difference of $+0.01$, the same multiple-object accuracy (92.3 vs.\ 92.5) and a
similar dynamic degree (74.2 vs.\ 74.7). \methodF{} runs at 12.5 FPS
($3.77\times$) with the same semantic score ($-0.03$). Cheaper single models do not reach
this trade-off: the 1.3B model loses 1.08 semantic points,
mostly in multiple objects (85.5), and two or three 14B steps lose 1.63 and 0.83
points at only $1.77\times$ and $1.31\times$. On the 797 original short prompts,
where the 1.3B model trails by 6.82 points, \method{} and \methodF{} close the
gap entirely ($-0.02$ and $-0.03$). Including VAE decoding, \method{} is
$2.34\times$ faster end to end; its first-chunk latency equals the 14B model's
(2.09\,s), and holding both models raises peak memory from 60.2 to 68.6\,GB.

\paragraph{Large first beats small first.}
Schemes that let the small model go first pay for verification. Exact
speculative decoding accepts 98.6\% of elements but no complete chunk-step and
ends up slower than the 14B model ($0.90\times$). SDVG verifies with a reward
model instead and matches the 14B model's semantics, but at 6.5 FPS;
\method{} reaches the same semantic score (82.18 vs.\ 82.16) $1.55\times$
faster (Fig.~\ref{fig:vbench_qual}). SRDiffusion-style cascades, which call the 14B model on every chunk,
lose 0.97 points with one 14B step ($1.80\times$) and need two steps
($1.40\times$) to close the gap: once the scene is set, per-chunk large-model
calls buy little.

\begin{figure}[t]
\centering
\includegraphics[width=\columnwidth]{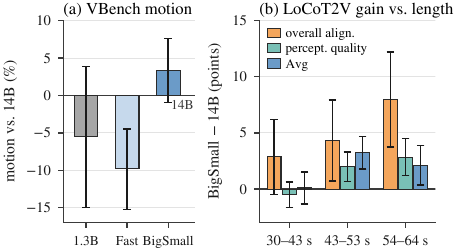}
\caption{(a) Motion on the VBench dynamic-degree prompts (mean RAFT peak flow,
relative to the 14B model). \methodF{} keeps the semantics but
loses motion; the sketches of \method{} restore it. (b) Gain of \method{} over
the 14B model on LoCoT2V-Bench by thirds of the video length: the
longer the video, the larger the gain.}
\label{fig:analysis}
\end{figure}

\paragraph{Two operating points and a division of labour.}
Without the plan chunk, the semantic score falls by 0.86 points
(Tab.~\ref{tab:main_results}). \methodF{}, which keeps only the plan chunk,
matches \method{} in semantics (paired difference $-0.05$)
but not in motion (Fig.~\ref{fig:analysis}a): on the dynamic-degree prompts, the mean RAFT peak flow drops
by 9.8\% of the 14B model's mean, and the sketches of
\method{} restore it ($+14.5\%$ over \methodF{};
vs.\ 14B: $+3.3\%$). \methodF{} is thus the choice when
speed matters most, and \method{} when the 14B model's motion must be kept. The plan chunk sets
\emph{what} is shown, sketches keep it \emph{moving}, and the small model
renders the rest.

\subsection{Minute-Long Generation}
\label{sec:locot2v}

\begin{figure}[t]
\centering
\includegraphics[width=\columnwidth]{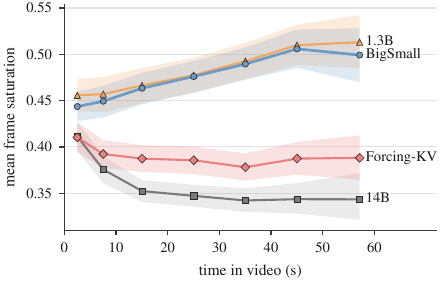}
\caption{\textbf{The 14B model fades over a minute.} Mean frame saturation over
time on LoCoT2V-Bench. The 1.3B model and \method{}
do not fade; Forcing-KV, the 14B model attending to a few recent frames,
fades much less than the 14B model.}
\label{fig:locot2v_drift}
\end{figure}

\paragraph{The 14B model loses its advantage.}
Over minute-long videos the 14B model is no better than the 1.3B model (65.83
vs.\ 65.72; Tab.~\ref{tab:locot2v}).
\method{} improves over both: $+1.78$ over the 14B model and $+1.43$ over the 1.3B model, at $3.14\times$ the
speed of the 14B model (8.6 vs.\ 2.7 FPS). The gain over the 14B model comes
from overall alignment ($+4.94$, fine-grained alignment
($+4.28$), character consistency ($+4.23$) and perceptual quality ($+1.39$),
and grows with the video length (Fig.~\ref{fig:analysis}b). \methodF{}, whose
only 14B computation is the first chunk, keeps the alignment gains (overall
$+4.16$, fine-grained $+5.57$ over the 14B model) but not the quality: its
average is not above the 14B model ($+0.77$), and the
sketches of \method{} add $+1.11$, mostly in PQ and HERD.
Over a minute, the first chunk still decides what is shown, and the planner is
needed again where the content changes.

\paragraph{Why.}
The 14B model degrades when it reads a long history of its own generations.
Its mean frame saturation falls from 0.41 to 0.35 within 20\,s and does not
recover (Fig.~\ref{fig:locot2v_drift}), and it often stays in its opening
scene while the prompt moves on, the same inertia it shows at prompt switches
(Sec.~\ref{sec:context}).
\method{} keeps the 14B model where it is strong: its planner sees only a sink
frame and the previous chunk, and the long-range state is carried by the
renderer, which
was trained on its own rollouts \cite{huang2025selfforcing} and does not fade.

\paragraph{Comparison with acceleration baselines.}
SDVG scores close to \method{} ($-0.33$), but it is slower
than the 14B model itself ($0.91\times$): over a minute only 38\% of its 1.3B
drafts pass the reward check, and every rejected chunk is regenerated by the
14B model after being decoded and scored. Letting the small model go first
thus becomes more expensive exactly where long videos need speed. Speculative
Forcing loses 4.15 points, mainly in PQ and HERD.
Forcing-KV scores highest, consistent with our diagnosis that the 14B model
suffers from its long context, but it is only $1.45\times$ faster than the 14B
model, as the 14B model still generates every chunk.

\paragraph{Prompt switches.}
\label{sec:prompt_switch}
On 172 videos whose prompt switches every seven chunks, \method{} scores
$+10.0$ over the 14B model on switched segments at
$3.22\times$ its speed, on par with the 1.3B model, with the smoothest
switches of all accelerated variants.

\subsection{Generality}
\label{sec:generality}
\begin{table}[t]
\centering
\caption{\textbf{Other renderers} (VBench, Krea-14B planner, thresholds
not re-calibrated). Gap: semantic gap to the 14B model closed relative to the
renderer alone.}
\label{tab:renderers}
\setlength{\tabcolsep}{2.5pt}
\resizebox{\columnwidth}{!}{
\begin{tabular}{llcccc}
\toprule
Renderer & Method & $\Delta$Sem (95\% CI) & Gap & M-obj. & Dyn. \\
\midrule
\multirow{3}{*}{Self-Forcing} & alone & $-2.51$ [$-4.33,-0.67$] & 0 & 86.4 & 65.3 \\
 & \methodF{} & $-0.47$ [$-1.08,+0.10$] & 0.81 & 90.4 & 72.2 \\
 & \method{} & $-0.18$ [$-0.75,+0.35$] & 0.93 & 91.1 & 70.8 \\
\midrule
\multirow{3}{*}{LongLive} & alone & $-2.30$ [$-4.20,-0.38$] & 0 & 85.1 & 40.3 \\
 & \methodF{} & $-0.34$ [$-1.04,+0.34$] & 0.85 & 90.8 & 61.1 \\
 & \method{} & $-0.41$ [$-1.06,+0.20$] & 0.82 & 91.2 & 63.9 \\
\midrule
\multirow{3}{*}{CausVid} & alone & $-5.87$ [$-7.92,-3.88$] & 0 & 84.2 & 66.7 \\
 & \methodF{} & $-1.35$ [$-2.20,-0.57$] & 0.77 & 88.9 & 77.8 \\
 & \method{} & $-1.18$ [$-1.93,-0.51$] & 0.80 & 89.7 & 75.0 \\
\bottomrule
\end{tabular}}
\end{table}

\paragraph{Other renderers.}
The order effect is not specific to Self-Forcing. We replace the renderer by two
independently trained 1.3B AR models of the same family, LongLive
\cite{yang2025longlive} and CausVid \cite{yin2025causvid}, without re-calibrating
any threshold (Tab.~\ref{tab:renderers}). In every case, the 14B first chunk
closes about 80\% or more of the semantic gap: with LongLive, \methodF{} and
\method{} are indistinguishable from the 14B model, and with CausVid, the weakest
renderer ($-5.87$ alone), they gain 4.5 to 4.7 points but remain below the 14B
model.

\subsection{Ablation Studies}
\label{sec:ablation}
\begin{table}[t]
\centering
\caption{\textbf{Ablations} (VBench, seed 0, 944 videos). $^\ast$The 869 prompts
of Tab.~\ref{tab:renderers}. $^\S$A 14B step 1 on every chunk.}
\label{tab:ablation}
\setlength{\tabcolsep}{3pt}
\resizebox{\columnwidth}{!}{
\begin{tabular}{llcccc}
\toprule
 & Variant & Speed-up & $\Delta$Sem (95\% CI) & Dyn. & M-obj. \\
\midrule
\multicolumn{2}{l}{\textbf{\method{}}} & $\mathbf{3.02\times}$ & $\mathbf{-0.12}$ [$\mathbf{-0.7,+0.4}$] & \textbf{75.0} & \textbf{91.2} \\
\midrule
\multirow{2}{*}{Plan chunk} & none & $3.98\times$ & $-2.31$ [$-3.9,-0.8$] & 65.3 & 84.5 \\
 & 2 of 4 steps by 14B & $3.68\times$ & $-0.76$ [$-2.0,+0.5$] & 72.2 & 88.0 \\
\midrule
\multirow{4}{*}{Sketches} & no run cap & $3.28\times$ & $-0.13$ [$-0.7,+0.4$] & 65.3 & 90.7 \\
 & skip target 70\%$^\ast$ & $3.01\times$ & $-0.18$ [$-0.8,+0.4$] & 70.8 & 91.1 \\
 & skip target 50\%$^\ast$ & $2.80\times$ & $+0.13$ [$-0.4,+0.6$] & 72.2 & 91.2 \\
 & skip target 30\%$^\ast$ & $2.67\times$ & $+0.12$ [$-0.4,+0.6$] & 75.0 & 91.3 \\
\midrule
\multirow{2}{*}{KV cache$^\S$} & exact, sparse & $2.02\times$ & $-0.07$ [$-0.6,+0.4$] & 72.2 & 91.8 \\
 & + window, late exit & $2.34\times$ & $-0.07$ [$-0.6,+0.4$] & 76.4 & 91.2 \\
\bottomrule
\end{tabular}}
\end{table}

\emph{The plan chunk carries the semantics} (Tab.~\ref{tab:ablation}): without it, the semantic score
falls to the level of the 1.3B model ($-2.31$ vs.\ $-2.50$), and with only two
of its four steps by the 14B model, a third of the gap remains ($-0.76$).
\emph{Sketches carry the motion}: without the run cap, the dynamic degree
falls to that of the 1.3B model (65.3); capping consecutive skips at two
restores it (75.0). Lower skip targets only cost speed, so $\tau$ needs no fine tuning. \emph{The approximate cache is free}:
the windowed planner cache together with late exit is 16\% faster than an
exact sparse cache at an identical semantic score.

\section{Conclusion}
\label{sec:conclusion}
Where in an AR video rollout is a large model needed? Where content is
introduced. In few-step AR video diffusion, the
model that generates the first chunk decides what the video shows, and later
chunks largely inherit it; giving the 14B model the first chunk recovers 95\%
of its semantic advantage at 9\% of its additional cost, whereas giving it every
later chunk recovers only 16\%. Letting the small model go first is therefore the
wrong order. The small model only has to carry the content forward, and over a minute it does so more stably than the 14B model, whose
quality degrades as it reads its own long history.
\method{} turns this division of labour into a training-free planner--renderer
sampler: the 14B model plans the opening chunk and sketches content changes,
the 1.3B model renders the rest, and a cross-model KV cache lets the planner
leave and re-enter the rollout at low cost. On VBench, \method{} is $3.02\times$ faster than the 14B
model without measurable semantic loss, and \methodF{} reaches $3.77\times$; on
minute-long videos, \method{} is both faster ($3.14\times$) and better than the
14B model. The order effect holds for three independently trained renderers,
suggesting that the question is not only how large a model to use, but when.

{
    \small
    \bibliographystyle{ieeenat_fullname}
    \bibliography{main}
}

\end{document}